\documentclass[manuscript,screen,review=false]{acmart}

\AtBeginDocument{%
  \providecommand\BibTeX{{%
    \normalfont B\kern-0.5em{\scshape i\kern-0.25em b}\kern-0.8em\TeX}}}

\setcopyright{acmcopyright}
\copyrightyear{2022}
\acmYear{2022}
\acmDOI{XXXXXXX.XXXXXXX}

\acmConference[Conference acronym 'XX]{Special Issue on Trustworthy AI}{
  2023}{Woodstock, NY}
\acmPrice{15.00}
\acmISBN{978-1-4503-XXXX-X/18/06}

\begin{document}

\title[Secure Robotics]{Secure Robotics: A Definition and a Brief Review from a Cybersecurity Control and Implementation Methodology Perspective}

\author{Adam Haskard}
\email{adam.haskard@bluerydge.com}
\affiliation{
  \institution{Bluerydge Cybersecurity and Technology}
  \streetaddress{P.O. Box 1212}
  \city{Canberra}
  \state{ACT}
  \country{AUS}
  \postcode{43017-6221}
}
\author{Damith Herath}
\orcid{1234-5678-9012}
\email{damith.herath@canberra.edu.au}
\affiliation{Collaborative Robotics Lab,
  \institution{University of Canberra}
  \streetaddress{P.O. Box 1212}
  \city{Canberra}
  \state{ACT}
  \country{AUS}
  \postcode{43017-6221}
}
\author{Zena Assaad}
\orcid{}
\email{zena.assaad@anu.edu.au}
\affiliation{
  \institution{The Australian National University}
  \streetaddress{}
  \city{Canberra}
  \state{ACT}
  \country{AUS}
  \postcode{}
}


\begin{abstract}
 Secure robotics is a multi-disciplinary endeavour for improving the cybersecurity posture of robotic and embodied Artificial Intelligence systems. The article surveys emerging concepts and ideas encapsulating the notion of secure robotics and identifies five Secure Robotics Cybersecurity Control Implementation Layers as a crucial starting point for consideration by practitioners. It also recognises the need for further studies on the relationship between Human-robot trust and the implementation of established and novel cybersecurity controls.
  
\end{abstract}



\maketitle

\section{Introduction}

With the proliferation of Information Technologies (IT) comes the risk of attacks on critical systems and sensitive information. As expected, recent high-profile and significant breaches have eroded trust in these systems. \textit{Cybersecurity}~\cite{cavelty2010cyber} is an umbrella term that defines several domains that work together to provide strategies and technologies to protect such systems and data from being compromised. The field of cybersecurity has matured over the years to counter these increasingly prevalent threats.  

However, little to no attention has been placed on potentially similar vulnerabilities and trust issues in robotics. Particularly with many robotic and other embodied and embedded Artificial Intelligence (AI) systems coming online in the wild with little human oversight, the potential risks have significantly increased in recent times. In this context, '\textit{Secure Robotics}' defines an umbrella term that would parallel cybersecurity within the IT domain to capture the techniques and strategies to secure vulnerable robotic systems from potential harm to (re)establish trust in robots and embodied AI systems.

This paper is a survey of the 'Secure Robotics' literature and of how the implementation of cybersecurity controls into a robotic system deployment may increase human-robot trust and robot system robustness amongst the robot user community. Further, this paper discusses the relationship between trust and cybersecurity as a vehicle for human-robot trust development and to further establish the role of cybsersecurity and trust within the secure robotics domain.

The impetus behind increased robotic deployment is broadly described in the Industry 4.0~\cite{busby2022social, lasi2014industry} concept.  Industry 4.0 has been expedited through a series of technological advancements, and a key element to technological progression is the development of new forms of human and machine interactions. Developments of note include augmented-reality systems, and systems which better utilise touch interfaces and other hands-free operating systems, a further contributing factor is the innovations in bandwidth to transfer digital data to physically usable machinery. Resultant examples include the improvements in advanced robotics and rapid prototyping. \cite{Gilchrist16} Driven by 'Industry 4.0' factors, the proliferation of robotic systems into societal domains provides opportunities to explore human-robot trust.

As a necessary factor of constructive relationships, including relationships between human participants and robotic systems, \cite{Martelaro16} improving trust factors can be considered an element of interest within secure robotics. In this context, noting several definitions of trust in Human-Robot Interaction (HRI) have been proffered including \begin{quote}“the reliance by one agent that actions prejudicial to the well-being of that agent will not be undertaken by influential others”\end{quote} \cite{Oleson11} and \begin{quote} "a belief, held by the trustor, that the trustee will act in a manner that mitigates the trustor’s risk in a situation in which the trustor has put its outcomes at risk”.\end{quote} \cite{Wagner11}. These definitions share a common question arouund, “whether a robot’s actions and behaviors correspond to human’s interest or not?” \cite{Zahra11} et al. Secure robotics viewed through the lens of the application of cybersecurity controls could consider its impacts on the human-robot trust relationship, as measuring the cybersecurity to trust relationship may become a critical element in the design and development of robotic systems. 

Events generated by human-robotic system interaction can provide data for the broader cultural impacts of robotic system deployments, including how they relate to trust. Secure robotics research provides research opportunities to assist in categorising the effects of the human-robotic system trust event stimulus. Measuring robotic system robustness with data produced from robotic system cybersecurity event perspective may provide an opportunity to better understand the fundamental nature of the relationship between secure robotics, trust, robustness and the robotic system participant. Researching the complex nature of this relationship could explore the contributions of secure robotics to the human-robot trust relationships.

In considering broad ranging factors contained within the human-robot trust relationship and the cultural impacts of secure robotics, invoking Asimov's laws as an established robotic system cultural artefact adds another dimension to the nature of the described secure robotics relationships. Derived originally from science fiction author Isaac Asimov in 1942, adherence to the three laws ensures the human participant is the primary beneficiary. The secure robotics relationship with robotic system participants trust in an Industry 4.0 context, presents more opportunities for Asimovs laws of robotics to be tested. 

At present, robotic systems are frequently deployed in an unsecured configuration vulnerable to cyber attack. Unsecured system configurations become particularly vulnerable when connected with wide area networks, public internet infrastructure or aggregated information and communication systems. \cite{Chen-Ching09} Unsecured robotic and autonomous systems may test the shared commonality of the previously described definitions, in that in a cyber compromised state the robotic system may not illicit behaviours which correspond to the humans interest. Presently, secure robotics remains an under-studied problem within the broader robotics discipline and further the affects of secure robotics on system robustness and human-robot trust. The advent of more frequently deployed robotic systems, exp[oses the vulnerabilities of these system to malicious acts to, on, or with robots. \cite{Lofaro16} 

The status of robot cybersecurity has recently been reviewed over a three year period, providing a description on the inherently unsecured nature of deployed robotic systems.
\begin{quote} "Some reasons for unsecured robotic deployment practices have been described as threefold:
first, defensive security mechanisms for robots are still on their early stages, not covering the complete threat landscape. Second, the inherent complexity of robotic systems makes their protection costly, both technically and economically. Third, vendors do not generally take responsibility in a timely manner, extending the zero-days exposure window (time until mitigation of a zero-day) to several years on average. Worse, several manufacturers keep forwarding the problem to the end-users of these machines or discarding it." \cite{Mayoral-Vilches21} \end {quote} 

Improving trust factors, engineering optimisation opportunities and the cybersecurity posture of deployed robotic systems through the implementation of established and novel cybersecurity controls is the defining feature of a secure robotic deployment. As a result of the survey, this paper proposes a five layer secure robotic control model and explores the need for more research into the genesis and application of cybersecurity controls specifically in the HRI layer. 

\section{Legacy Robotics: Vulnerable production systems interfacing with high threat environments}
The cybersecurity posture of Supervisory Control and Data Acquisition (SCADA) Systems is of increased interest, due to the ever increasing connectivity of critical infrastructure such as power grids with information and communication systems\cite{Cheng22}. Similar to SCADA systems, robotic systems face the threat of cyber-attacks when interfacing with high threat environments. Moreover, once the robotic system weaknesses become well known its likely threat actors will seek out the vulnerable systems for exploitation. Legacy robotic systems are susceptible on local networks. Local networks refer to ways for unwanted users to access and/or affect the operation of the Remote Internet of Things (RIoT) device when on the local network. The local network includes the communication between the primary control computer and the actuators. Examples of this is the CAN (Controller Area Network). This is one of the biggest security concerns for robotics. \cite{Lofaro16}. Local network vulnerabilities are then amplified in likelihood of execution when connected to a Wide Area Network. More research needs to be done to ascertain if legacy robotic system with limited to null cybersecurity rigour applied lower the human-robot trust relationship.

\section{Secure Robotics: The implementation of cybersecurity controls to the robotic environment}
The continued viability and growth of economic sectors ensures that companies are motivated to design and build cybersecurity elements into the architecture of their products and systems. \cite{Krizz11}  . 
Secure Robotics considers the implementation of established cybersecurity controls to robotic environments. The Australian Signals Directorate (ASD), an Australian Government cybersecurity agency, has established fundamental 'cybersecurity Principles' \cite{ASD22}. The purpose of the cybersecurity principles is to provide strategic guidance on the protection of systems and data from cyber threats. 
These high level principles provide the overarching framework in the survey of cybersecurity control implementation. The cybersecurity principles are grouped into four key activities; govern, protect, detect and respond. The subsections below are a short survey of the principles in a secure robotics context.

\subsection{Secure Robotics: Governance}
A key element of the govern principle relevant to secure robotics is that security risks are identified, documented, managed and accepted both before systems and applications are authorised for use, and continuously throughout their operational life. Comprehensive security risk assessments of robotic systems which analyses configuration data, logging materials and embedded application weaknesses provide the descriptive baseline risk landscape for robotic system owners and users. Outputs from robotic system security assessments generate data sets which indicate the key areas of concern in any given deployment. Generated data sets also identify global risks with a particular deployment. Global robotic system risks include operating system, embedded application and physical access vectors \cite{Lofaro16} all of which requires security control implementation and continuous monitoring to address. In a comprehensive overview of challenges involved in intelligent connectivity in IoT-assisted robotics by Dutta et al. 
\cite{Dutta21} found that an application of cybersecurity standards and a robotics software framework is therefore needed. More research needs to be done to develop comprehensive secure robotic governance frameworks.

\subsection{Secure Robotics: Protection}
 An element of the protect principle, security vulnerabilities in systems and applications are identified and mitigated in a timely manner is of particular relevance to robotic systems. Vulnerability data affecting a vendors robots shows how according to historical data, vulnerabilities were patched as early as 14 days after its disclosure however the average mitigation time is above four years (1500 days) \cite{Mayoral-Vilches21}. Reduction in mitigation time can be contributed to by personnel who are provided with ongoing cybersecurity awareness training. Security education training awareness controls include awareness programs for changing organisational attitudes to realise the importance of security and the adverse consequences of its failure, training which provides the relevant skills to personnel to enable them to perform their security activities more effectively, and specific education which is targeted for IT security professionals and focuses on developing the ability and vision to perform complex, multi-disciplinary activities. \cite{NIST98} Cybersecurity, Education, Training and Awareness (SETA) in a robotic system context is a complex multi-disciplinary endeavour for existing cybersecurity professionals as there is limited publicly available literature on the matter, thus requiring more research. The limited secure robotics training data available may contribute to the proliferation of unsecured systems however more research is required to better understand the secure robotics and training relationship.

\subsection{Secure Robotics: Detection}
 Dutta et al.\cite{Dutta21} concludes that the detection of cyber malfeasance needs to be considered when designing a modern robotic system that uses IoT resources and that in real-time, it is paramount to understand that that the infusion to the control systems of malicious malware and/or data launches hazards, may have severe consequences. To treat the risks of cyber threats, different detection techniques are available. These techniques are typically classified as either network-based or host-based.\cite{Li16}  Opportunities for both techniques are relevant for secure robotics considering key underlying detection principles. \cite{ASD22} 
\begin{itemize} 
\item  Event logs are collected and analysed in a timely manner and
\item Cybersecurity events are analysed in a timely manner to identify cybersecurity incidents. 
\end{itemize}
Robotic software frameworks such as Robot Operating System (ROS)~\cite{quigley2009ros,st2022robot} or Robot Operating System 2 (ROS2) produce logs and events during run-time. Log collection and analysis is a keystone activity for a security incident and event monitoring solution. Secure Robotics considers the implementation of detection principles to establish a baseline cybersecurity posture, offer engineering data and research opportunities for the human-robot trust relationship.  More research is required to ascertain the benefits of implementing a human in the loop for security incident and event management activities within a robotic deployment. While there is an economic benefit in removing the human entirely from automated systems, there is a realisation that it is important to include the abilities of humans as part of the industrial process. The problem is that complex systems are optimised for automatic control, and in many cases this includes robotics, hence minimal human intervention is necessary. However, broader human intervention is beneficial if the robotic system develops faults, or if there is a window of opportunity which long term optimisation misses but the human with excellent pattern perception can notice.  \cite{Moray94}
The implementation of principles  may provide cybersecurity, trust and optimisation benefits necessary underpinning Secure Robotics, however given the limited literature available on the topic more research is required.  
\subsubsection{human-robot trust and engineering optimisation through security incident and event management}
Cybersecurity security control families designed for gaze and image recognition systems and their effects on human-robot trust are currently undeveloped requiring more research. This research opportunity potentially includes feasibility studies into stenographic based 'attack patterns' being used to deny system availability or to achieve administrator access of a deployed a system. Security incident and event management (SIEM) requires the collection and analysis of security events or incidents within a real-time information processing environment. The SIEM provides a centralised view of the security scenario of an IT infrastructure. In the context of a robotic system, a deployed SIEM can ingest robotic and autonomous system logs from its parent operating system. The logs can be tuned to contain engineering data which is useful for later optimisation activities. Big data analytic techniques from collected data sets provide additional opportunities for engineering optimisation, the opportunities multiple if a Machine Learning or Artificial Intelligence capability provides big data analytical support to identify patterns, trends and cyber malfeasance.

\subsection{Secure Robotics: Response}
In the cybersecurity knowledge domain, the term 'Incident analysis' references a detailed process which examines security data and evaluates incident causal factors and data relationships. The analysis phase, requires incident response personnel to identify and develop indicators of compromise (IOC) that the system owner and its delegates can use to drive containment and eradication activities.
Incident response personnel must possess advanced skills and experience in the area of digital forensics for a comprehensive analysis to be completed. \cite{Gorecki20} Depending on characteristics and variables of an incident, analysis activities may include the following:
\begin{itemize} 
\item  Acquiring and preserving forensic data
\item  Developing investigative leads
\item  Analyzing host and network data
\item  Malware analysis
\item  Cyber Threat Intelligence enrichment
\end {itemize}

Secure Robotics considers incident response principles as relevant in their application through a typical analysis activity. Three key incident response principles are, 
\begin{itemize} 
\item Cybersecurity incidents are reported both internally and externally to relevant bodies in a timely manner
\item Cybersecurity incidents are contained, eradicated and recovered from in a timely manner and 
\item Business continuity and disaster recovery plans are enacted when required.\cite{ASD22} 
\end {itemize}
Due to the limited research on robotic system incident response currently available, an opportunity is identified for research literature to be developed addressing: 
\begin{itemize}
    \item Incident response in a robotic system context
    \item Robotic system logging and monitoring infrastructure
    \item Robotic system incident response hardware implementations
    \end{itemize}  
Opportunities exist for novel robotic system incident response research including robotic system hardware circuits designed to ship data to a cloud based security and incident and event management platform.    Robotic system incident response remains under-studied. 

\section {Secure Robotics Cybersecurity Control Implementation Layers}
\begin{figure}[h]
  \centering
  \includegraphics[width=\linewidth]{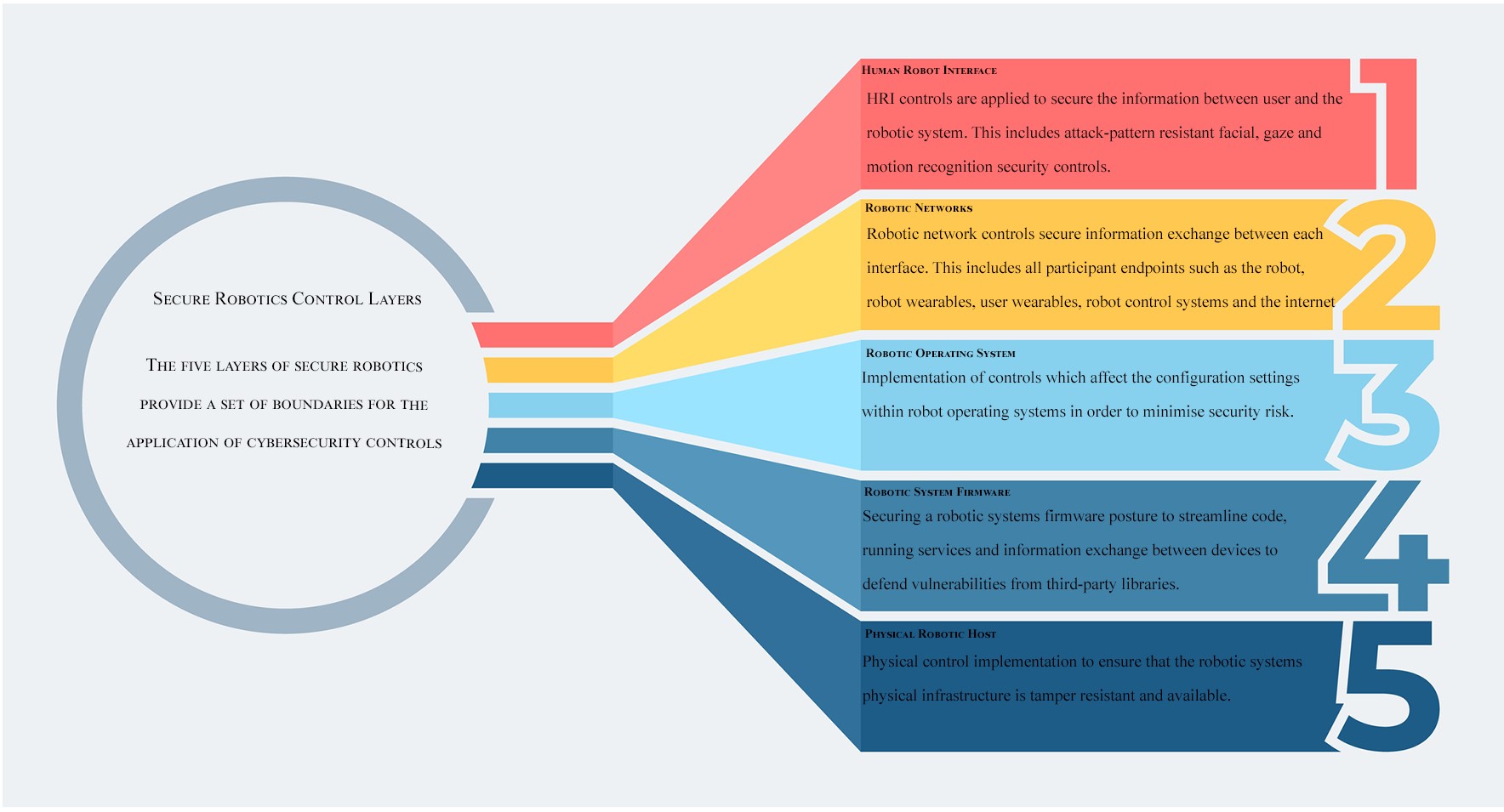}
  \caption{Secure Robotics Control Layers}
  \label{fig: Secure Robotics Control Layers}
  \Description{Secure Robotics Control Layers.}
\end{figure}

This paper surveyed security layers for robotic systems, including the Human-Robot Interface, Robotic Networks, Robot Operating System, Robotic System Firmware and the Physical Robotic Host as demonstrated in Fig~\ref {fig: Secure Robotics Control Layers}. The secure robotics control layers in  Fig~\ref {fig: Secure Robotics Control Layers} can be considered complementary to the targeting layers classification provided in \cite{Yaacoub21}. To ground the control implementation layers, a review of several emerging qualitative and quantitative robotic risk assessment methodologies was conducted by J.P A. Yaacoub et. el. in \cite{Yaacoub21}, the team identified relevant qualitative assessments for robotics including Operationally Critical Threat, Asset, and Vulnerability Evaluation (OCTAVE) which is useful for evaluating risks based on a risk acceptance level without focusing on risk avoidance and “Méthode Harmonisée d’Analyse de Risque (MEHARI)" which ensures a quantitative risk assessment of risk components, and is based on measuring the maturity of system level. \cite{Yaacoub21} further identifies an absence of difficult to achieve quantitative risk assessments in the robotic field, and presents their own Robotic Risk Assessment (RRS) method which is based on the evaluating the likelihood and impact of an attack. 

 For the purposes of this paper we propose a five layer security control implementation concept for robotic systems which considers the implementation of cybersecurity controls into and across each information transaction interface to treat risks identified in an RRS assessment. We propose that the implementation of cybersecurity controls into each layer may increase human-robot trust factors.  However, cybersecurity control implementation requires a balance to be achieved between technical, economic and sustainment feasibility and that the balance treats the risks. Additionally, cybersecurity controls must be designed and implemented appropriately to ensure access and operation to legitimate activities and system functions remain uninhibited, particularly during periods of human safety criticality or restoration activity particularly pertinent for robotic system functionality. \cite{NIST20} This section surveys the existing literature describing the application of cybersecurity control families through the lens of the five layers of 'Secure Robotics' and their effects on human-robot trust factors.

\subsection{Secure Robotics Control Layer: Physical Robotic Hardware}
Prone to hardware based attack and being complex amalgamations of software and hardware designed to function materially in space. \cite{Yaacoub21} presents many hardware protection solutions from the available literature. The NIST Physical and Environment Protection control family may address typical physical system vulnerabilities applied to the robotic system hardware setting, however this survey identified that more research is required for robotic system specific physical protection weaknesses and their effects on human-robot trust factors. 

\subsection{Secure Robotics Control Layer: Robotic System Firmware}
 \cite{Yaacoub21} et. al. notes that it is essential for robotic system firmware to be updated and points to opportunities for a standardised OS and authentication process. In addition to standardisation, sound secure coding practices like those in the Secure Software Development Framework (SSDF) would need to be added to and integrated with each Software Development Life-cycle (SDLC) implementation as software development life cycle (SDLC) models would need to explicitly address robot software security in detail. Further research needs to be conducted to ascertain if robotic firmware can support the implementation of NIST control SI-7(15), Code Authentication, which is where Cryptographic authentication can verify if software or firmware elements have been digitally signed using approved and recognised digital certificates and the implementation effects on human-robot trust relationships. However \cite{Teixeira20} notes that cryptography can compromise the dependent real-time robotic system performance. More research is required to identity if unintended changes to firmware via an attack in a robotic system may increase risks to human safety and erode human-robot trust factors.

\subsection{Secure Robotics Control Layer: Robot Operating System}
In \cite{Teixeira20} \textit{Security on ROS: analyzing and exploiting vulnerabilities of ROS-based
systems} a broad analysis of related work is conducted and reference is made to aligning the three defined pillars of cybersecurity, Confidentiality, Integrity and Availability (CIA) to make a robotic system secure, further a method is presented which is capable of of compromising the CIA on a ROS based system. Finally, the experiment demonstrated that it was possible for a malicious user to compromise the confidentiality, integrity and availability of the information processing in a robotic system. As \cite{Mayoral-Vilches21} states, continued offensive security activities need to be undertaken against robotic systems. As a further research opportunity the findings from each threat emulation event can be used to develop new security controls, and the implementation of which can be measured in the context of the human-robot trust relationship.

\subsection{Secure Robotics Control Layer: Robotic Networks}
Yaacoub et al.~\cite{Yaacoub21} provides a comprehensive review on in the implementation of existing and novel cryptographic solutions and protocols including Transport Layer Security (TLS) and Datagram
Transport Layer Security (DTLS) in the ROS core to secure the robot communication. This solution provides a fine-grained control over permissions to publish, subscribe or consume data. In addition a robotics communication security control is presented in \cite{Dutta21}. 
Communication security control is a critical component secure robotics, and should be considered foundational in future research as the implementation of these solutions relate to human-robot trust factors.

\subsection{Secure Robotics Control Layer: Human Robot Interaction}
While trust has been defined within the context of HRI~\cite{hancock2011meta}, as discussed in section 1 of this paper, this term is reflective of a subjective concept that can be difficult to quantify. The essence of trust is that of integrity - a measure of the trust which can be placed in the correctness of the information supplied by the total system \cite{ICAO02} The integrity of information is a core pillar of cybersecurity. Quantifying or measuring the level of trust a human may hold in a robotic system is difficult to capture and requires further investigation. While accuracy of information or data can be measured, other attributes of define human-robot trust, such as risk mitigation \cite{Wagner11} – for varying types of risk - are more complex to quantify. 

\section{Conclusion }

Secure robotics is a multi-disciplinary endeavour for improving the cybersecurity posture, human-robot trust and engineering optimisation opportunities of robotic systems through the implementation of established and novel cybersecurity controls. Opportunities for research identified in this survey are several. In particular, understanding the risks of legacy robotic systems interfacing with high-threat environments, similar to the cybersecurity posture of Supervisory Control and Data Acquisition (SCADA) Systems interfacing with information and communication systems and understanding how human-robot trust is affected by the implementation of cybersecurity controls become apparent. The paper further surveyed the implementation of cybersecurity controls in the robotic environment in the context of the ASD cybersecurity principles of governance, protection, detection, and response. Each principal provides areas of consideration for robotic system cybersecurity solution implementation. This paper further provided a novel conceptual five-layer security control implementation model for deployment of secure robotic systems. The layers for consideration are the Human-Robot Interface, Robotic Networks, Robot Operating System, Robotic System Firmware and Physical Robotic Host (Fig~\ref {fig: Secure Robotics Control Layers}). This survey also identified that although a handful of robotic systems have been developed with  robotic security controls, their effects on human-robot trust is not well understood and the human-robot Interface in the cybersecurity control and human-robot trust context remains largely understudied.

\bibliographystyle{ACM-Reference-Format}
\bibliography{sample-base}

\end{document}